\documentclass[lettersize,journal]{IEEEtran}
\usepackage{amsmath,amssymb,amsfonts}
\usepackage{algorithmic}
\usepackage{algorithm}
\usepackage{array}
\usepackage[caption=false,font=normalsize,labelfont=sf,textfont=sf]{subfig}
\usepackage{textcomp}
\usepackage{stfloats}
\usepackage{url}
\usepackage{verbatim}
\usepackage{graphicx}
\usepackage{cite}
\usepackage{xcolor}
\usepackage{booktabs}
\usepackage{multirow}
\usepackage{bm}
\begin{document}

\title{Geometry-Aware Fisheye-LiDAR Fusion for Robust 3D Object Detection in Low-Overlap Setups}

\author{
Xiangzhong Liu,
Xihao Wang,
and Hao Shen%
\thanks{
Xiangzhong Liu, Xihao Wang, and Hao Shen are with 
Technical University of Munich, 80333 Munich, Germany
(e-mail: xiangzhong.liu@tum.de).
}
}

% The paper headers
% \markboth{IEEE Robotics and Automation Letters}%
% {Anonymous Authors: Geometry-Aware Fisheye-LiDAR Fusion}

\maketitle

\begin{abstract}
As autonomous systems expand from capital-intensive robotaxis to cost-sensitive
logistics, sensor configurations are increasingly optimized for coverage-per-cost.
A prevalent sparse-view setup utilizes omnidirectional dual-fisheye cameras
with a roof-mounted LiDAR, introducing severe geometric challenges: extreme
radial distortion, minimal overlap, and misalignment between spherical projections
and rectilinear grids. BEV fusion algorithms typically force image and
point cloud modalities into unified Cartesian or polar grids early in
the pipeline, causing significant feature distortion and information loss
for wide-view fisheye cameras. To address this, we propose a Geometry-Aware
Hybrid Fusion (GA-HF) framework that explicitly accounts for fisheye geometry
and BEV feature distortion, where fisheye features are lifted into a
polar BEV grid via a Distortion-Aware Lift-Splat-Shoot (LSS) module to
preserve native angular density, while LiDAR features are processed in native
Cartesian space for metric fidelity of bounding box regression. To
bridge these heterogeneous streams, we introduce a Dual-Attention Warping
Correction module that applies spatial and channel attention to the warped
camera features before fusion, explicitly suppressing artifacts in low-quality
peripheral regions while enhancing high-quality semantic cues. GA-HF is evaluated
on three benchmarks: KITTI-360, Dur360BEV, and the synthetic Fisheye3DOD datasets.
To the best of our knowledge, it is the first approach to explore LiDAR–fisheye camera fusion.
On KITTI-360, GA-HF improves NDS by 4.2\% over Cartesian baselines; on Dur360BEV, it surpasses
both LiDAR-only and BEVFusion, while significantly reducing orientation error
despite the geometric distortions; and on Fisheye3DOD, it attains the
highest detection score among all fusion methods. These results validate the
necessity and generalizability of geometry-aware fusion for sparse sensor setups.
\end{abstract}

\begin{IEEEkeywords}
Sensor Fusion, Fisheye Cameras, 3D Object Detection,
Bird's-Eye View, Attention Mechanisms
\end{IEEEkeywords}

\section{Introduction}
\label{sec:intro}

\IEEEPARstart{T}{he} commercial deployment of autonomous ground vehicles (AGVs) in logistics,
mining, and last-mile delivery requires perception systems that strike a balance
between redundancy and hardware cost. While Level 4 robotaxi fleets utilize
dense sensor suites, e.g., 6+ overlapping pinhole cameras and multiple
LiDARs, to create a seamless 360$^{\circ}$ perception field, cost-sensitive platforms
often adopt sparse configurations. A typical setup involves a single top-mounted
LiDAR and dual fisheye cameras, a configuration mirrored in multiple real-world
datasets, such as KITTI-360 \cite{liao2022kitti}, Dur360BEV \cite{wenke2025dur360bev},
and Oxford Spires \cite{tao2025spires}, among others.

This sparse multi-view setup introduces unique geometric challenges. The
Cartesian BEV representation of standard fusion frameworks~\cite{liu2022bevfusion,bai2022transfusion,yan2023cross}
fundamentally mismatches the non-uniform distribution of image information,
where nearby regions possess significantly denser visual evidence yet
receive the same spatial resolution as distant ones. While this inefficiency
is tolerable in dense camera setups, it becomes a critical bottleneck in
sparse-view settings where each wide-FOV camera must cover vast areas with
limited redundancy. Recent works like PolarFormer \cite{jiang2023polarformer}
and PolarBEVDet \cite{yu2024polarbevdet} adopt polar BEV grids, which better
align sampling resolution with the radial distribution of image data to
mitigate perspective-induced information imbalance.

\begin{figure}[!t]
\centering
\includegraphics[width=\columnwidth]{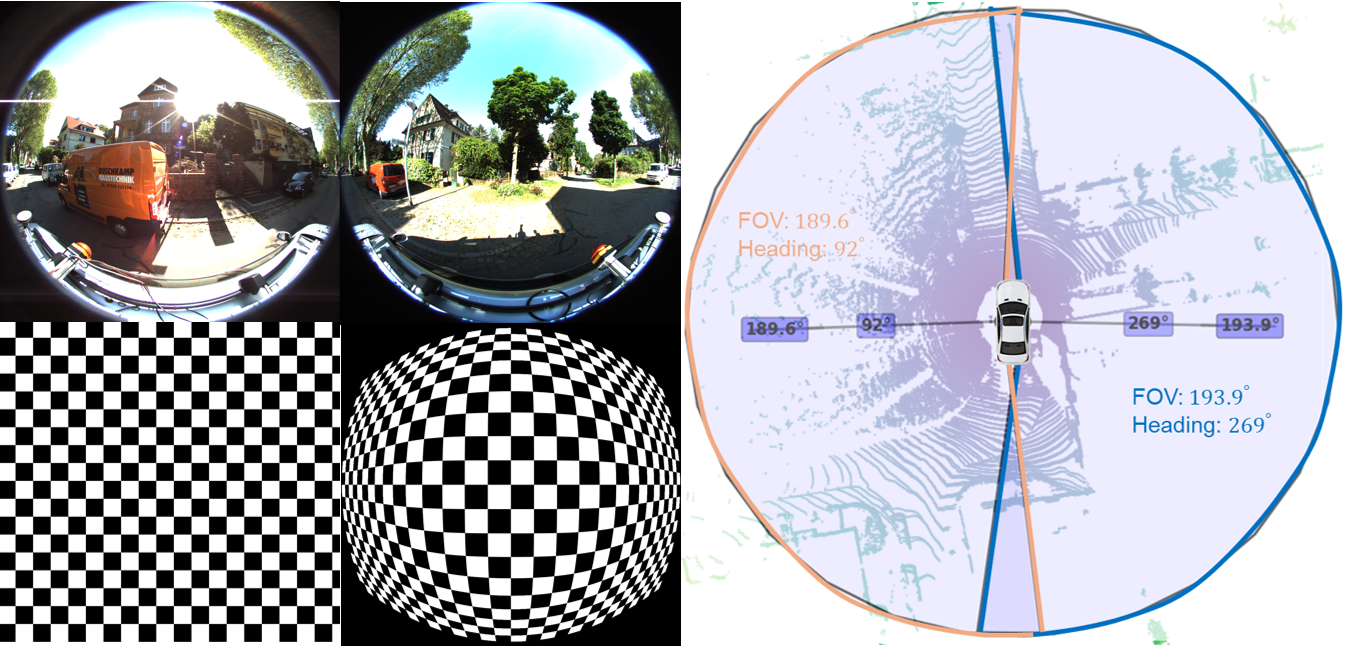}
\caption{Sparse sensor configuration on KITTI-360. The lateral dual
fisheye cameras (orange and blue sectors) cover $\approx 190^{\circ}$
FOV each, leaving minimal overlap in the front and rear. Fisheye lenses
introduce severe radial distortion compared to pinhole cameras, resulting
in non-uniform angular resolution that compresses features at the periphery.}
\label{fig:sensor_setup}
\end{figure}

In the sparse sensor setting, this motivation is further amplified due to
severe radial distortion of fisheye cameras, where non-uniformity arises not
only along the radial dimension but also across viewing angles. Fisheye
projections exhibit highly non-uniform angular pixel sampling, with angularly
dense observations near the optical center but compressed and heavily warped
toward the image periphery. Consequently, a polar BEV representation aligns naturally
with the fisheye sensor's angular geometry, ensuring high-fidelity feature
preservation during the lifting process.

However, operating purely in the polar domain introduces intrinsic challenges
for object detection. As highlighted in LiDAR-based studies such as PARTNER
\cite{nie2023partner} and PolarStream \cite{chen2021polarstream}, rigid objects
with rectangular shapes become severely distorted and rotation-variant in polar
coordinates, appearing as angle- and range-dependent warped shapes. This
geometric inconsistency significantly complicates bounding box regression, forcing
prior LiDAR-centric works to rely on complex re-alignment modules and geometry-aware
adaptations to compensate for polar grid deformations.

To resolve this dilemma without introducing such architectural complexity,
we propose a Geometry-Aware Hybrid Fusion (GA-HF) framework. The key insight
is to leverage the strengths of both coordinate representations by decoupling
feature extraction from object detection: fisheye features are processed in
the polar domain to maximize angular fidelity, while multi-modal fusion and bounding
box regression are performed in the Cartesian field. To bridge these
representations effectively, we incorporate a Dual-Attention fusion mechanism
based on the CBAM \cite{woo2018cbam}. This module adaptively weights the
warped camera features, suppressing artifacts in high-distortion or low-overlap
regions before they are fused with the LiDAR stream. This hybrid design
leverages the inherent advantages of both coordinate systems, ensuring high-fidelity
feature preservation alongside robust, translation-invariant detection. Our main
contributions are:
\begin{enumerate}
    \item \textbf{Distortion-Aware Polar LSS.} We introduce a polar lifting
        scheme for fisheye cameras that projects image features into a polar
        BEV grid using the MEI fisheye camera model, preserving the native angular
        resolution and radial density of fisheye lenses.

    \item \textbf{Hybrid Coordinate Architecture.} We propose a dual-stream
        framework that decouples polar camera features from Cartesian LiDAR
        features. By performing object regression in the Cartesian domain,
        we ensure translation-invariant detection and avoid the geometric
        instabilities common in pure-polar detection heads.

    \item \textbf{Dual-Attention Warping Correction Module.} We integrate a
        Dual-Attention Warping Correction Module based on CBAM to resolve
        the cross-modal quality imbalance. The module learns a spatial reliability
        map to adaptively suppress camera artifacts in high-distortion and low-overlap
        regions.

    \item \textbf{Cross-Dataset Validation.} We evaluate GA-HF on three
        benchmarks: the real-world
        KITTI-360 and Dur360BEV datasets and the synthetic Fisheye3DOD
        benchmark, demonstrating consistent improvements across diverse
        domains and class distributions.
\end{enumerate}

\section{Related Work}

\subsection{BEV Representations for 3D Object Detection}
Contemporary 3D object detection has largely converged on the BEV representation
to provide a unified space for multi-sensor fusion. Cartesian-based methods,
such as BEVFusion \cite{liu2022bevfusion} and TransFusion \cite{bai2022transfusion},
project features onto uniform grids, which struggle with the non-uniform information
density of radial lenses but are still tolerable for dense surrounding camera
configurations. To address this, PolarFormer \cite{jiang2023polarformer} and
PolarBEVDet \cite{yu2024polarbevdet} introduced polar grids to pinhole camera
models to better align with the depth-dependent distribution of image pixels.
In the LiDAR segmentation domain, works like PC-BEV \cite{qiu2025pc} and Cylinder3d
\cite{zhou2020cylinder3d} explore cylindrical or polar partitions to handle the
inherent sparsity of point clouds. However, as noted in PARTNER \cite{nie2023partner}
and PolarStream \cite{chen2021polarstream}, performing bounding box
regression directly in polar coordinates introduces significant geometric
instability due to rotation variance and shape distortion. Consequently,
these polar-based methods require extra complex feature re-alignment modules
and geometry-aware adaptations to compensate for the feature distortions
induced by polar coordinate transformations, adding computational overhead and
architectural complexity. Beyond single-modal methods, PolarFusion \cite{shi2025polarfusion} 
unifies LiDAR and camera features in a polar coordinate system to reduce misalignment and 
enable effective BEV fusion. PolarGFusion3D \cite{li2024polargfusion3d} extends this idea 
with graph attention over polar structures to fuse LiDAR, camera, and radar 
data while alleviating distortion and inefficiency. OccCylindrical \cite{ming2025occcylindrical} 
instead operates in cylindrical coordinates, improving geometric fidelity for 3D semantic occupancy 
prediction. However, these fully polar-centric architectures often
inherit the regression difficulties of polar grids and are primarily designed
for multi-view pinhole cameras. As the first framework to fuse fisheye
cameras and LiDAR, our GA-HF framework strategically leverages polar coordinates
only for feature lifting, while transitioning to Cartesian coordinates for
detection. This approach avoids the regression instabilities of purely polar-based
methods while preserving metric consistency.

\subsection{Fisheye Camera for 3D Perception}
Traditional fisheye processing relies on rectification to pinhole models,
which introduces artifacts and discards wide-FOV information at the image
periphery. Distortion-aware 2D methods such as RectConv
\cite{griffiths2024adapting}, DarSwin \cite{athwale2023darswin}, and Calibrated
Convolutions \cite{Berenguel-Baeta_2023_BMVC} demonstrate that incorporating
fisheye models improves feature preservation. However, extending to 3D
perception presents distinct challenges that exceed the scope of per-view
distortion modeling, which fails to guarantee consistent cross-camera feature
extraction required by multi-view object detection. Recent works target BEV segmentation with surrounding-view cameras
(FishBEV \cite{li2025fishbev}, FisheyeBEVSeg
\cite{yogamani2024fisheyebevseg}), while 3D detection methods such as F2BEV
\cite{samani2023f2bev} and Fisheye3DOD~\cite{li2025exploring} inject fisheye
projection models into view transformation modules. While these works 
establish strong baselines for fisheye 3D detection and segmentation, they 
remain confined to camera-only, synthetic evaluation and do not address 
the challenges of LiDAR fusion under sparse sensor configurations.
FisheyeDepth \cite{zhao2025fisheyedepth} conducted depth estimation with distortion modeling on real-world data. EquivFisheye~\cite{yang2025equivfisheye} uses a rotation-equivariant spherical framework 
to fix distortion and improve consistency, though spherical convolutions increase computational costs. Our work
bridges this gap by utilizing a Polar LSS module specifically designed for fisheye
geometry and fusing with LiDAR data.

\begin{figure*}[!t]
\centering
\includegraphics[width=\textwidth]{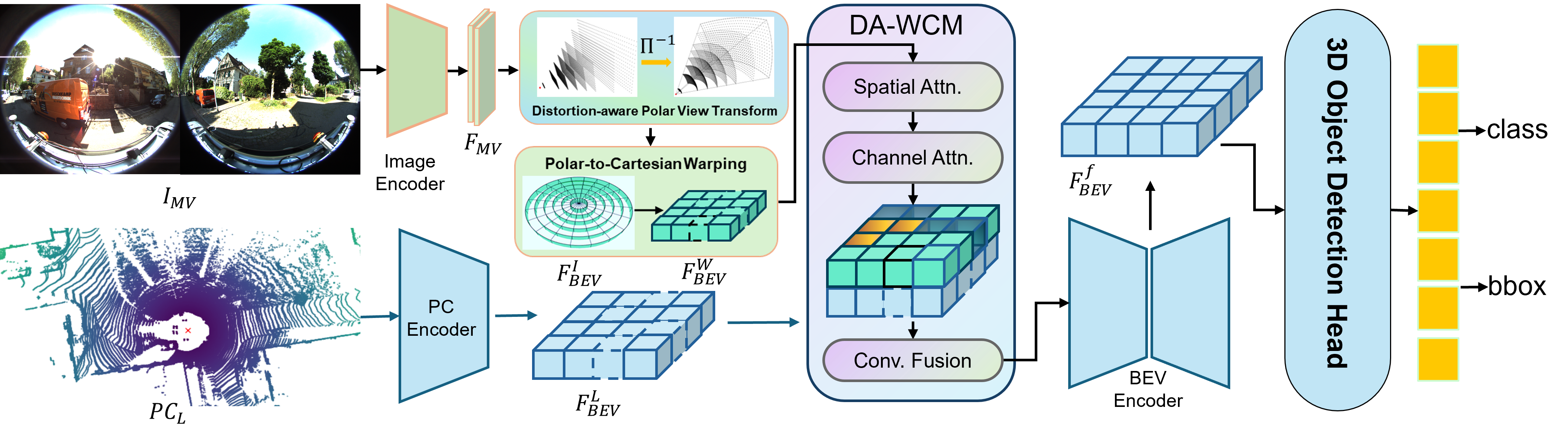}
\caption{Overview of the proposed Geometry-Aware Hybrid Fusion (GA-HF)
framework for 3D object detection. Visual features are lifted to polar BEV
grid using the MEI camera model ($P_{MEI}$) to handle distortion, then
warped and refined via DA-WCM to fuse with Cartesian LiDAR features
derived directly from point cloud.}
\label{fig:arch}
\end{figure*}

\subsection{Cross-Modal Attention Fusion}
Effective fusion requires managing the feature quality imbalance among different
sensor modalities. Early BEV-based fusion strategies, such as BEVFusion \cite{liu2022bevfusion},
adopt straightforward feature concatenation or summation, implicitly assuming
comparable reliability across modalities. TransFusion
\cite{bai2022transfusion} and CMT \cite{yan2023cross} utilize transformer-based
cross-attention to selectively aggregate cross-modal features. DAL \cite{huang2024detecting}
emphasizes the importance of modality-aware reliability modeling over symmetric
fusion, proposing that camera features contribute only to classification and
confidence estimation while geometric regression remains grounded in LiDAR's
metric precision. To further refine feature reliability, SElayer
\cite{hu2018squeeze} and CBAM \cite{woo2018cbam} have been employed to
provide channel and spatial weighting. In the context of fisheye-LiDAR fusion,
this imbalance is amplified by extreme radial distortion and non-uniform
sensor overlap. While DAL can manage this by dropping camera features for regression
tasks, it simply discards the necessary dense visual and semantic cues in
wide-FOV settings. We propose an alternative DA-WCM that deploys CBAM-based weighting
to explicitly refine warped fisheye features, ensuring that the dense but
potentially noisy fisheye features do not degrade the high-precision LiDAR
representation.

\section{Methodology}

We propose the Geometry-Aware Hybrid Fusion (GA-HF) framework, which
decouples feature extraction from object detection. As illustrated in
Fig.~\ref{fig:arch}, it consists of two parallel streams: a polar fisheye
stream and a Cartesian LiDAR stream, bridged by the DA-WCM for feature fusion in the
Cartesian domain, with CenterPoint or
TransFusion~\cite{bai2022transfusion} heads for 3D regression.

 \subsection{Distortion-Aware Polar LSS}
    Standard LSS assumes a pinhole camera model, lifting features via uniform
    depth bins along $Z$. This fails for fisheye lenses with non-linear
    distortion and non-uniform angular resolution. We formulate a
    model-agnostic polar lifting module that abstracts the camera-specific
    projection into a generic unprojection interface $\Pi^{-1}$, decoupling
    geometric back-projection from downstream polar BEV construction and
    enabling seamless adaptation across heterogeneous fisheye camera models.

    Instead of a Cartesian frustum, we define a cylindrical frustum $\mathcal{V}_{polar}
    \in \mathbb{R}^{R \times \Theta \times Z}$ in the vehicle ego coordinate,
    where $R$ and $\Theta$ denote radial distance and azimuth, respectively.

    \textbf{Generic Unprojection Interface.}
    The lifting operation requires an accurate unprojection from pixel
    coordinates $\mathbf{u}=(u, v)^{\top}$ to unit-norm 3D ray directions
    $\mathbf{X}' \in \mathbb{S}^{2}$:
    \begin{equation}
        \mathbf{X}' = \Pi^{-1}(\mathbf{u};\,\boldsymbol{\theta}_{calib})
        \label{eq:generic_unproj}
    \end{equation}
    where $\boldsymbol{\theta}_{calib}$ denotes camera-specific calibration
    parameters. Given predicted depth bins $D$, the 3D point in the camera
    frame is $\mathbf{P}_{cam}= D \cdot \mathbf{X}'$, transformed to the LiDAR
    frame via extrinsics. The unprojection map is precomputed and cached per
    camera, so the ray lookup table is computed once and reused across all
    frames sharing the same intrinsics.
    We instantiate $\Pi^{-1}$ for the fisheye models used in this work.

    \textbf{(i) MEI Unified Model} (KITTI-360 \cite{liao2022kitti}).
    Pixel coordinates are first converted to distorted normalized coordinates
    $x_{d}= (u - u_{0})/\gamma_{1}$, $y_{d}= (v - v_{0})/\gamma_{2}$,
    where $\gamma_{1,2}$ are focal lengths and $(u_{0}, v_{0})$ is the principal
    point. The undistorted radius $r$ is recovered by inverting the radial
    distortion model $x_{d}= x(1 + k_{1}r^{2}+ k_{2}r^{4})$ via
    Newton-Raphson iteration:
    \begin{equation}
        r_{n+1}= r_{n}- \frac{r_{n}(1 + k_{1}r_{n}^{2}+ k_{2}r_{n}^{4}) - \sqrt{x_d^2
        + y_d^2}}{1 + 3k_{1}r_{n}^{2}+ 5k_{2}r_{n}^{4}}
    \end{equation}
    The undistorted coordinates $(x, y) = (x_{d}, y_{d}) / (1 + k_{1}r^{2}+ k_{2}r^{4})$
    are then back-projected onto the unit sphere following the MEI model
    \cite{mei2007single}:
    \begin{equation}
        \begin{aligned}
            Z' & = \frac{\xi + \sqrt{1 + (1 - \xi^{2})(x^{2}+ y^{2})}}{x^{2}+ y^{2}+ 1}, \\
            X' & = x\,(Z' + \xi),\quad Y' = y\,(Z' + \xi).
        \end{aligned}
    \end{equation}
    where $\xi$ is the mirror parameter.

    \textbf{(ii) Polynomial Fisheye Model} (Fisheye3DOD
    \cite{li2025exploring} and Dur360BEV \cite{wenke2025dur360bev}).
    Fisheye3DOD uses a polynomial angular fisheye projection, and for
    Dur360BEV we fit the same OCam-style polynomial family to the dual-fisheye
    imagery for a unified implementation. For the OCam unprojection, image coordinates 
    are first corrected by the affine parameters and principal point to obtain $(x_{p}, y_{p})$, from which the
    image radius is computed as $r=\sqrt{x_{p}^{2}+y_{p}^{2}}$. The axial
    component is then recovered by the direct polynomial
    \begin{equation}
        z_{p}= -\sum_{i=0}^{4} a_{i}r^{i}
        \label{eq:ocam_proj}
    \end{equation}
    where $\{a_{i}\}_{i=0}^{4}$ are calibration coefficients. The 3D ray is
    obtained by normalizing the lifted point,
    $\mathbf{X}'= [y_{p},\,x_{p},\,z_{p}]^{\top}/\|[y_{p},x_{p},z_{p}]\|_{2}$.
    Because Dur360BEV provides a single dual-fisheye image covering $360^{\circ}$,
    we split the frame into front and rear hemispheres and apply the unprojection
    independently for each before mapping the rays to the ego frame.

    \textbf{Polar BEV Construction.}
    Regardless of the camera model, once $\mathbf{X}'$ is obtained, a
    Cartesian-to-polar mapping populates the cylindrical BEV
    grid $(\rho, \phi)$ as $\rho = \sqrt{X^{2}+ Y^{2}}$ and
    $\phi = \arctan(Y/X)$. The mapping normalizes the fisheye lens's
    non-linear distortion into a regular grid with linear radial and angular
    relationships, avoiding interpolation artifacts of Cartesian resampling.

\subsection{Hybrid Coordinate Alignment}
While the polar fisheye stream $\mathbf{F}_{polar}$ preserves the sensor's
native angular fidelity, the LiDAR stream provides absolute metric precision
essential for bounding box regression. The LiDAR point cloud is processed by a
sparse voxel encoder (VoxelNet) to produce Cartesian features
$\mathbf{F}_{lidar}\in \mathbb{R}^{H \times W \times C}$. Since
$\mathbf{F}_{polar}$ and $\mathbf{F}_{lidar}$ reside in topologically distinct
manifolds, we perform differentiable coordinate warping for spatial alignment.

A sampling grid $\mathcal{G}_{xy}$ in the Cartesian LiDAR feature space is
converted to polar coordinates via $\mathcal{M}$ with normalization to polar
BEV ranges. The warped fisheye feature is obtained by bilinear interpolation:
\begin{equation}
    \mathbf{F}_{fish}= \text{GridSample}(\mathbf{F}_{polar}, \mathcal{G}_{xy
    \to \rho\phi})
\end{equation}
While this aligns modalities in Euclidean space, the resulting
$\mathbf{F}_{fish}$ contains anisotropic stretching and noise in regions
corresponding to the fisheye periphery or low-overlap sectors, necessitating
targeted refinement before fusion.

\subsection{Dual-Attention Warping Correction Module}
The DA-WCM resolves the inherent quality imbalance between the high-precision
LiDAR stream and the distortion-prone fisheye stream. In sparse-view
configurations, polar-to-Cartesian warping induces non-linear stretching
artifacts in low-overlap regions. The DA-WCM utilizes a modified CBAM
\cite{woo2018cbam} to explicitly refine the warped visual features before
integration with the LiDAR stream.

\textbf{Spatial Reliability Modeling:} Given the warped camera feature $\mathbf{F}
_{fish}\in \mathbb{R}^{H \times W \times C_{cam}}$, we compute a spatial
reliability map by aggregating channel-wise information:
\begin{equation}
    \mathbf{M}_{s}(\cdot) = \sigma\left(f^{7\times7}\left(\left[\operatorname{AvgPool}
    _{c}(\cdot); \operatorname{MaxPool}_{c}(\cdot)\right]\right)\right)
\end{equation}
where $f^{7\times7}$ is a $7\times7$ convolution.
This map $\mathbf{M}_{s}\in [0, 1]^{H \times W}$ adaptively downweights
unreliable regions, including blind sectors and areas affected by severe
peripheral distortion.

\textbf{Channel-Wise Feature Selection:} A channel attention module with
shared MLP (reduction ratio $r=16$) computes $\mathbf{M}_{c}\in
\mathbb{R}^{C}$:
\begin{equation}
    \mathbf{M}_{c}(\cdot) = \sigma\!\left( \operatorname{MLP}(\operatorname{AvgPool}
    _{s}(\cdot)) + \operatorname{MLP}(\operatorname{MaxPool}_{s}(\cdot)) \right
    )
\end{equation}

These weights are applied exclusively to the warped camera stream to
generate a corrected feature map $\mathbf{F}_{cam}^{*}$:
\begin{equation}
    \mathbf{F}_{cam}^{*}= \mathbf{M}_{c}\otimes (\mathbf{M}_{s}\otimes \mathbf{F}
    _{warp})
\end{equation}
The corrected camera features are concatenated with $\mathbf{F}_{lidar}$ and
passed through a convolution-based fusion block to produce the final BEV
feature $\mathbf{F}_{fused}$. This asymmetric refinement preserves LiDAR
metric integrity while the camera stream contributes high-density semantic
context only where deemed reliable by the attention modules.
\begin{figure}[!t]
\centering
\includegraphics[width=\columnwidth]{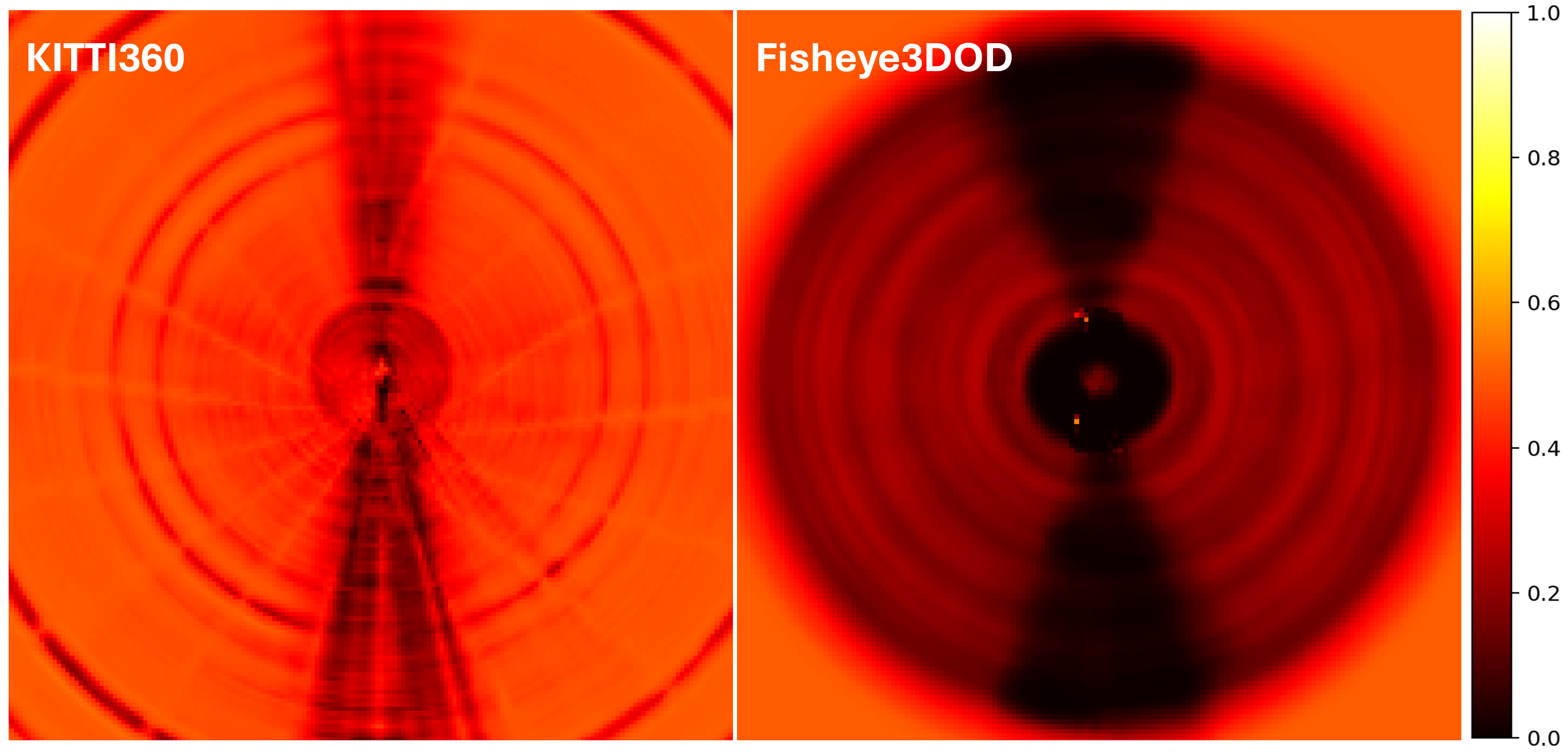}
\caption{Visualization of the spatial attention map
    $\mathbf{M}_{s}$ averaged over 50 samples on KITTI-360 and Fisheye3DOD, with 
    left-right fisheye cameras: the stable black
    fan-shape in front and rear area, proving DA-WCM learns a scene-independent
    geometric prior purely from data.}
\label{fig:attn_weight}
\end{figure}

\section{Experiments}

\subsection{Datasets and Setup}
We evaluate our framework on three benchmarks that span real-world and
synthetic domains with distinct fisheye-LiDAR configurations.

\textbf{KITTI-360} \cite{liao2022kitti} provides a Velodyne HDL-64E LiDAR
and dual lateral fisheye cameras with MEI calibration. Following the nuScenes-format conversion
pipeline from~\cite{liu2026benchmarking}, we evaluate 10 classes, \textit{car, truck, trailer, bus, bicycle, motorcycle, pedestrian, pole,
object, and traffic sign},
across $\sim$11k uniformly sampled training frames (from 56k total, reducing
10\,Hz temporal redundancy) and 8,500 full validation frames.

\textbf{Fisheye3DOD} \cite{li2025exploring} is a synthetic benchmark generated
via the CARLA simulator, providing four surround-view fisheye cameras
($\text{FoV}>220^{\circ}$) and LiDAR with 3D bounding box annotations across
6 classes: \textit{bus, car, cyclist, pedestrian, truck}, and \textit{van}.
As a synthetic dataset with controlled conditions, it serves as a complementary
testbed to verify cross-domain generalizability.

\textbf{Dur360BEV} \cite{wenke2025dur360bev} is a real-world dataset with a
dual-fisheye camera (front and rear) and roof-mounted LiDAR, repurposed from BEV
segmentation for 3D object detection across 3 classes: \textit{car, cyclist}, and
\textit{pedestrian}. Notably, sensor calibration parameters for camera model
and extrinsics are not officially provided and were estimated through
calibration fitting, which may limit the achievable camera-LiDAR alignment
quality for fusion methods on this benchmark.

\textbf{Evaluation Protocol.}
Since KITTI-360 lacks attribute annotations, NDS is computed excluding
mAAE, following~\cite{liu2026benchmarking}. For cross-dataset comparison we
additionally report NDS$_{-v}$, which further excludes mAVE following the
Fisheye Detection Score (FDS) protocol of Fisheye3DOD~\cite{li2025exploring}:
\begin{equation}
    \mathrm{NDS}_{-v} = \dfrac{1}{6} \Big[3\,\mathrm{mAP}
    + \!\!\sum_{\mathrm{mTP} \in \mathbb{TP}_{-v}}\!\! \left( 1 - \min \left( 1, \mathrm{mTP} \right) \right) \Big]
    \label{eq:nds_no_mave}
\end{equation}
where $\mathbb{TP}_{-v} = \{\mathrm{mATE}, \mathrm{mASE}, \mathrm{mAOE}\}$.
NDS$_{-v}$ is the primary metric on Fisheye3DOD and Dur360BEV (which lack
velocity annotations), while KITTI-360 as our primary benchmark retains both NDS and mAVE for
completeness.

\subsection{Implementation Details}
We implement GA-HF using the MMDetection3D~\cite{mmdet3d2020} codebase. The image backbone is
ResNet-50 initialized with ImageNet weights and a VoxelNet-style architecture
for the LiDAR stream. Fisheye images are pre-cropped to 256$\times$704 (KITTI-360), 384$\times$800 (Fisheye3DOD)
and 320$\times$640 (Dur360BEV) to
exclude irrelevant regions (ego-vehicle body and sky), eliminating the need for
explicit validity masking during splatting. We train for 20 epochs with AdamW
(lr=$1\times10^{-4}$, weight decay 0.01) on 2 NVIDIA A5000 GPUs with batch
size 16, without CBGS. To minimize inference
latency, we cache the polar frustum geometry during testing with consistent
camera intrinsics, bypassing the online Newton-Raphson iteration in unprojection. Our GA-HF
adds negligible complexity (35.8M parameters) compared to the BEVFusion baseline
(35.6M), ensuring efficient deployment.

\subsection{Main Results}

\begin{table*}[!t]
\centering
\caption{Evaluation on KITTI-360 validation set. NDS excludes mAAE
(unavailable); NDS$_{-v}$ further excludes mAVE
following~\cite{li2025exploring}. Bold: best; underline: second best.
($^{*}$: our re-implementation,
$^{\dag}$: trained with full 56k data frames, $(C)$: CenterPoint head, $(T)$:
TransFusion head)}
\label{tab:main_results_no_mave}
\begin{tabular}{l|cc|ccccccc|cc}
    \toprule[2pt] \textbf{Model}               & \textbf{Modality} & \textbf{Coord.} & \textbf{mAP}$\uparrow$ & \textbf{NDS}$\uparrow$  & \textbf{NDS}$_{-v}$ $\uparrow$ & \textbf{mATE}$\downarrow$ & \textbf{mASE}$\downarrow$ & \textbf{mAOE}$\downarrow$ & \textbf{mAVE}$\downarrow$ & \textbf{AP}$_{\text{car}}$ $\uparrow$  & \textbf{AP}$_{\text{bus}}$ $\uparrow$ \\
    \midrule BEVDet                            & F                 & cart.           & 0.035                  & 0.056                   & 0.069                                  & 0.945                     & 0.745                     & 1.127                     & 1.366                     & 0.192                                  & 0.0                                   \\
    DAP-BEVDet (ours)                          & F                 & polar           & 0.096                  & 0.160                   & 0.191                                  & 0.735                     & 0.406                     & 1.118                     & 0.960                     & 0.427                                  & 0.033                                 \\
    \midrule CenterPoint                       & L                 & cart.           & 0.417                  & 0.404                   & 0.469                                  & 0.333                     & 0.340                     & \textbf{0.764}            & 1.150                     & 0.760                                  & 0.166                                 \\
    TransFusion-L \cite{bai2022transfusion}    & L                 & cart.           & \underline{0.454}      & 0.424                   & \underline{0.487}                                  & \underline{0.308}         & \textbf{0.323}            & 0.811                     & 0.979                     & 0.770                                  & \underline{0.185}                     \\
    Cylinder3d \cite{zhou2020cylinder3d}$^{*}$ & L                 & polar           & 0.378                  & 0.396                   & 0.423                                  & 0.358                     & 0.342                     & 0.897                     & 0.744                     & 0.755                                  & 0.148                                 \\
    CMT-L \cite{yan2023cross}                  & L                 & cart.           & 0.364                  & 0.378                   & 0.406                                  & 0.395                     & 0.347                     & 0.911                     & 0.779                     & 0.750                                  & 0.096                                 \\
    \midrule BEVFusion(C)                      & L+F               & cart.           & 0.424                  & 0.405                   & 0.469                                  & 0.313                     & 0.342                     & 0.801                     & 1.173                     & \underline{0.785}                      & 0.179                                 \\
    PolarBEVFusion(C)$^{*}$                    & L+F               & polar           & 0.360                  & 0.397                   & 0.404                                  & 0.349                     & 0.350                     & 0.960                     & \underline{0.606}         & 0.766                                  & 0.148                                 \\
    CMT \cite{yan2023cross}                    & L+F               & cart.           & 0.368                  & 0.376                   & 0.413                                  & 0.408                     & 0.348                     & 0.871                     & 0.834                     & 0.767                                  & 0.058                                 \\
    DAL \cite{huang2024detecting}              & L+F               & cart.           & 0.402                  & 0.418                   & 0.418                                  & 0.329                     & 0.367                     & 1.018                     & \textbf{0.569}            & 0.778                                  & 0.165                                 \\
    \textbf{GA-HF (C)}                         & L+F               & hybrid          & 0.431                  & \underline{0.435}       & 0.476                                  & 0.309                     & 0.343                     & 0.785                     & 0.805                     & \textbf{0.793}                         & 0.183                                 \\
    \textbf{GA-HF (T)}                         & L+F               & hybrid          & \textbf{0.456}         & \textbf{0.447}          & \textbf{0.492}                      & \textbf{0.302}            & \underline{0.333}         & \underline{0.783}         & 0.829                     & 0.782                                  & \textbf{0.204}                        \\
    \midrule CenterPoint $\dag$                & L                 & cart.           & 0.419                  & 0.403                   & 0.458                                  & 0.308                     & 0.330                     & 0.872                     & 0.945                     & 0.785                                  & 0.207                                 \\
    GA-HF (T)$\dag$                            & L+F               & hybrid          & 0.470                  & 0.429                   & 0.493                                 & 0.241                     & 0.331                     & 0.878                     & 1.698                     & 0.824                                  & 0.171                                 \\
    \bottomrule[2pt]
\end{tabular}
\end{table*}
As shown in Table \ref{tab:main_results_no_mave}, our GA-HF framework
establishes a new state-of-the-art for fisheye-LiDAR fusion methods on the
KITTI-360 real-world dataset.

\textbf{Comparison with Single Modality:} Our DAP-BEVDet nearly triples
BEVDet's mAP and NDS, confirming that polar processing natively handles
fisheye distortion. Among LiDAR-only methods, TransFusion-L
\cite{bai2022transfusion} is the strongest baseline (0.454 mAP), though
LiDAR methods lack semantic richness for orientation estimation (mAOE: 0.811).

\textbf{Performance of Fusion Methods:} Existing fusion baselines struggle
to outperform LiDAR-only on KITTI-360 because standard projection introduces
feature artifacts in sparse-view setups. DAL achieves the best mAVE (0.569) by
excluding camera features from regression, but suffers the highest mAOE
(1.018) and sub-baseline mAP (0.402). Our GA-HF (C) significantly
outperforms DAL in mAP (0.431 vs.\ 0.402) and mAOE (0.785 vs.\ 1.018),
demonstrating that camera features are indispensable for orientation
estimation but require targeted refinement to avoid degrading regression
stability. GA-HF (T) achieves state-of-the-art performance (0.456 mAP, 0.447
NDS), notably outperforming TransFusion-L in AP$_{\text{bus}}$, where
semantic-geometric synergy is most critical.

Scaling to the full 56k dataset ($\dag$) boosts mAP to 0.470 and NDS$_{-v}$
to 0.493, though mAVE degrades (1.698) due to extreme class imbalance
(Cars: 430K vs.\ Bus+Tram: 1K) and annotation inconsistency destabilizing
velocity regression. As none of the compared methods employ temporal modeling,
velocity estimation remains inherently noisy across all single-frame
approaches.

\subsection{Cross-Dataset Evaluation}

To assess generalizability beyond KITTI-360, we evaluate GA-HF on two
additional benchmarks with distinct sensor configurations.
Table~\ref{tab:cross_dataset} summarizes the results.

\begin{table}[!t]
\centering
\caption{Cross-dataset evaluation on Fisheye3DOD (synthetic, left-right
fisheye cameras used + LiDAR, 6 classes) and Dur360BEV (real-world, front-rear fisheye
spherical camera + LiDAR, 3 classes). NDS$_{-v}$ follows Eq.~\ref{eq:nds_no_mave}.
Bold indicates the best.
$^{\ddag}$: using a ResNet-18 backbone and 4 fisheye cameras, consistent with~\cite{li2025exploring}.}
\label{tab:cross_dataset}
\begin{tabular}{l|l|ccc}
    \toprule[1pt]
    \textbf{Dataset} & \textbf{Model} & \textbf{mAP}$\uparrow$ & \textbf{NDS}$_{-v}$$\uparrow$ & \textbf{mAOE}$\downarrow$ \\
    \midrule
    \multirow{5}{*}{Fisheye3DOD}
    & FisheyeBEVDet$^{\ddag}$ (F)  & 0.382  & 0.485  & 0.480 \\
    & DAP-BEVDet$^{\ddag}$ (F)          & 0.459  & 0.540  & 0.441 \\
    & TransFusion-L (L)                 & 0.960  & 0.916  & 0.207 \\
    & BEVFusion (L+F)                & \textbf{0.967}  & 0.900  & 0.321 \\
    & \textbf{GA-HF} (L+F)          & 0.963  & \textbf{0.925}  & \textbf{0.178} \\
    \midrule
    \multirow{5}{*}{Dur360BEV}
    & BEVDet (F) & 0.173  & 0.201  & 0.981 \\
    & DAP-BEVDet (F)   & 0.191  & 0.216  & 0.986 \\
    & TransFusion-L (L)                 & 0.694  & 0.742  & 0.404 \\
    & BEVFusion (L+F)                & 0.599  & 0.573  & 1.022 \\
    & \textbf{GA-HF} (L+F)          & \textbf{0.706}  & \textbf{0.751}  & \textbf{0.354} \\
    \bottomrule[1pt]
\end{tabular}
\end{table}

\textbf{Fisheye3DOD.}
GA-HF achieves the highest NDS$_{-v}$ of 0.925, surpassing both TransFusion-L
(0.916) and BEVFusion (0.900), driven by a substantial mAOE reduction (0.178
vs.\ 0.321 for BEVFusion). Our camera-only DAP-BEVDet (0.540 NDS$_{-v}$) also
outperforms FisheyeBEVDet~\cite{li2025exploring} (0.485), validating the polar
lifting module independently. While all fusion methods achieve high mAP
($>$0.96) under synthetic conditions, BEVFusion exhibits the worst orientation
artifacts, reinforcing the motivation for our hybrid coordinate design.

\textbf{Dur360BEV.}
Under imperfect sensor calibration, BEVFusion fails catastrophically in orientation estimation (mAOE: 1.022, effectively
random orientation), causing its NDS$_{-v}$ (0.573) to fall far below the
LiDAR-only TransFusion-L baseline (0.742). In contrast, GA-HF surpasses
both LiDAR-only and BEVFusion performance, achieving the best mAP (0.706),
NDS$_{-v}$ (0.751), and mAOE (0.354). These results show that the hybrid
design and DA-WCM not only mitigate orientation degradation under imprecise
calibration, but also recover useful fusion gains on this challenging
real-world benchmark.

\subsection{Ablation Studies}

\subsubsection{Impact of Distortion-Aware Polar LSS}
Table \ref{tab:ablation_polar} shows that baseline BEVDet fails on fisheye
data due to rectilinear grid misalignment. Enabling distortion-aware (D.A.)
view transformation improves mAP by 71\%, while our full DAP-BEVDet (D.A. +
Polar) reaches 0.160 NDS, a two-fold increase over the baseline. This confirms
that processing fisheye features in polar space natively preserves geometric
density and mitigates perspective smearing. Doubling the input image
resolution (\textit{Large}) further boosts overall performance,
demonstrating that our polar representation scales effectively with higher
pixel density to capture finer semantic details in distorted regions.

\begin{table}[!t]
\centering
\caption{Ablation of the Distortion-Aware Polar LSS module (Camera-Only).}
\label{tab:ablation_polar}
\resizebox{\columnwidth}{!}{
\begin{tabular}{l|cc|ccc}
    \toprule \textbf{Model Config}       & \textbf{D.A.} & \textbf{Polar} & \textbf{mAP}$\uparrow$ & \textbf{NDS}$\uparrow$ & \textbf{AP}$_{\text{car}}$ $\uparrow$ \\
    \midrule BEVDet                      & -             & -              & 0.035                  & 0.056                  & 0.192                                 \\
    DA-BEVDet                            & \checkmark    & -              & 0.060                  & 0.096                  & 0.344                                 \\
    \textbf{DAP-BEVDet}                  & \checkmark    & \checkmark     & \textbf{0.096}         & \textbf{0.160}         & 0.427                                 \\
    \midrule \textit{DAP-BEVDet (Large)} & \checkmark    & \checkmark     & \textbf{0.139}         & \textbf{0.187}         & \textbf{0.490}                        \\
    \bottomrule
\end{tabular}
}
\end{table}

\subsubsection{Analysis of Fusion Mechanisms}
We compare DA-WCM against standard fusion strategies in Table
\ref{tab:ablation_fusion}. Compared to naive concatenation, channel
attention (SELayer) improves the NDS to 0.431 but fails to address spatial
misalignments. Cross-Attention (as used in TransFusion) improves metric scores
(0.435 NDS) but results in a lower mAP (0.418), likely due to the difficulty
of learning attention maps on sparse, distorted inputs. Our DA-WCM
achieves the best overall performance, indicating that explicitly modeling spatial
reliability and channel modulation are necessary to refine fisheye
distorted features to avoid corrupting LiDAR representations.

\begin{table}[!t]
\centering
\caption{Comparison of different fusion mechanisms within the GA-HF
framework (using CenterHead).}
\label{tab:ablation_fusion}
\resizebox{0.8\columnwidth}{!}{
\begin{tabular}{l|ccc}
    \toprule \textbf{Fusion Strategy} & \textbf{mAP}$\uparrow$ & \textbf{NDS}$\uparrow$ & \textbf{mAOE} $\downarrow$ \\
    \midrule Concat.                  & 0.424                  & 0.405                  & 0.801                      \\
    SELayer                           & 0.424                  & 0.431                  & 0.801                      \\
    Gated-Attn.                       & 0.426                  & 0.416                  & 0.823                      \\
    Cross-Attn.                       & 0.418                  & \textbf{0.435}         & 0.817                      \\
    \textbf{DA-WCM}                   & \textbf{0.431}         & \textbf{0.435}         & \textbf{0.785}             \\
    \bottomrule
\end{tabular}
}
\end{table}

\subsection{Angular Stratified Analysis}

\begin{table}[!t]
\centering
\caption{Angular-stratified NDS (\%) on KITTI-360 validation dataset.}
\label{tab:angular_nds}
\resizebox{\columnwidth}{!}{
\begin{tabular}{l|cccc}
    \toprule Method     & Right                 & Front                   & Left                  & Back                    \\
    \midrule DAP-BEVDet & 22.04                 & 14.32                   & 17.54                 & 16.29                   \\
    TransFusion-L       & 47.67                 & 46.01                   & 44.20                 & 41.92                   \\
    GA-HF               & 48.71(1.04$\uparrow$) & 45.97(0.04$\downarrow$) & 48.26(4.06$\uparrow$) & 41.49(0.43$\downarrow$) \\
    \bottomrule
\end{tabular}
}
\end{table}

Table \ref{tab:angular_nds} breaks down performance across different angular
sectors ($90^{\circ}$ each) regarding the coverage of dual side-mounted fisheye
cameras. GA-HF shows significant NDS gains in the lateral \textit{Left} (+4.06\%)
and \textit{Right} (+1.04\%) sectors where fisheye coverage is primary. Conversely,
in the \textit{Front} and \textit{Back} sectors, corresponding to the visual
blind spots or extreme periphery, GA-HF maintains LiDAR-level performance. This
validates that our attention mechanism successfully leverages visual semantics
where available while safely relying on LiDAR in sensor gaps, consistent
with the qualitative inspections of the learned attention maps shown in
Figure~\ref{fig:attn_weight}.

\subsection{Qualitative Results}
\begin{figure}[!t]
\centering
\includegraphics[width=\columnwidth]{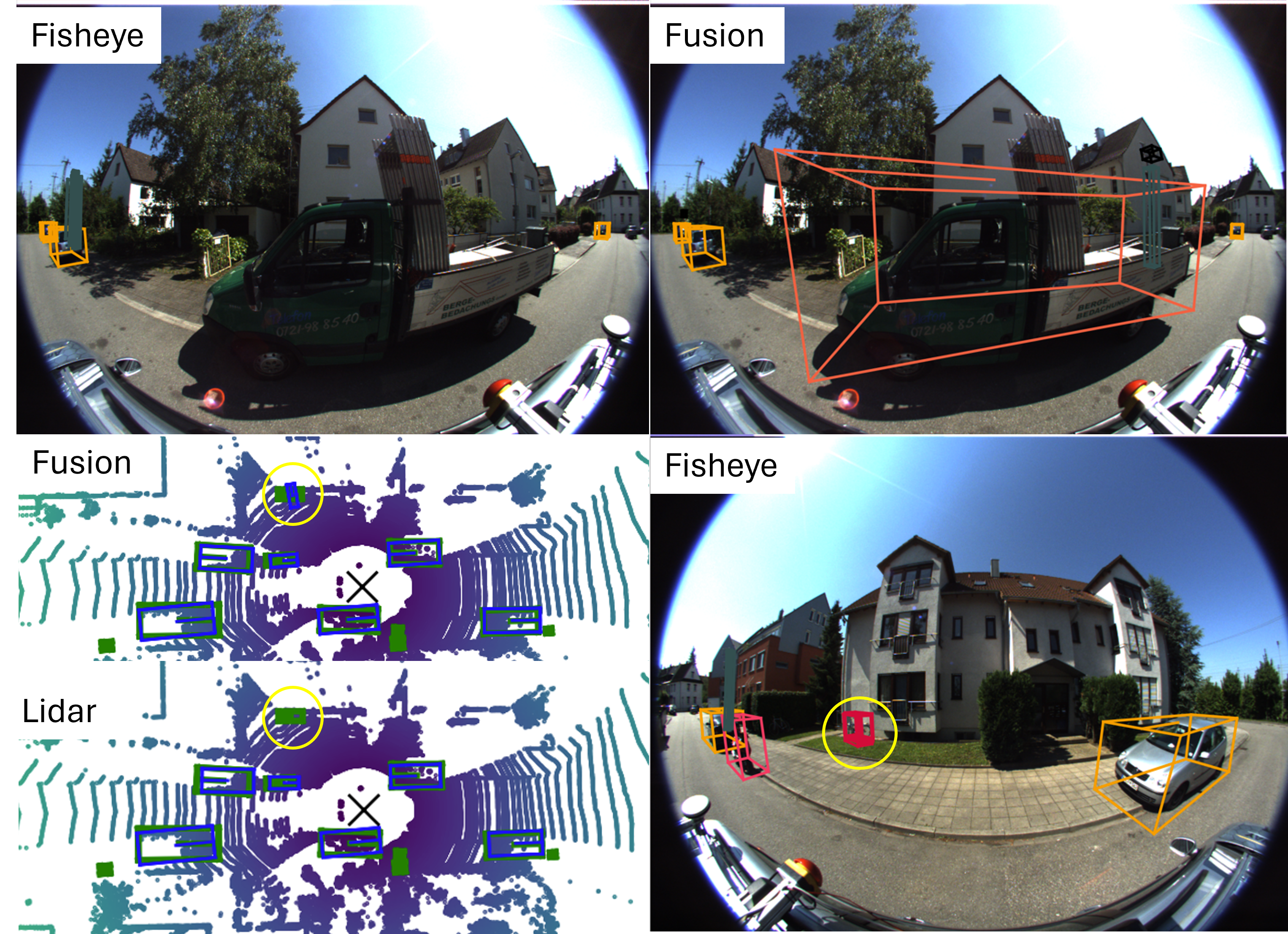}
\caption{Qualitative detection results on KITTI-360. Top row: the camera-only
model fails to detect a nearby truck, while GA-HF fusion correctly localizes
it with accurate 3D bounding boxes. Bottom row (BEV): a bicycle leaning
on a wall (yellow circle) is missed by LiDAR-only detection but is visible in the camera view and recovered
by GA-HF, which leverages dense fisheye semantic cues to complement the
sparse point cloud. Crosses mark the ego-vehicle position.}
\label{fig:quali_viz}
\end{figure}

Figure~\ref{fig:quali_viz} illustrates two representative failure modes of
single-modality detection that GA-HF resolves through geometry-aware fusion.
In the top row, a nearby truck occupying a large portion of the fisheye FOV is missed by
the camera-only model due to severe close-range distortion, yet GA-HF
correctly localizes it by anchoring to LiDAR geometry. In the bottom row,a bicycle
adjacent to a wall, which returns very thin, insufficient LiDAR points mixed entirely with the wall point cloud,
is missed entirely in LiDAR-only detection but recovered through
complementary fisheye evidence, with DA-WCM ensuring reliable integration
without degrading the fused representation. These
cases confirm that single-modality approaches are fundamentally limited in
sparse sensor setups and that targeted multi-modal fusion is essential for
robust perception across diverse scene geometries.

\section{Conclusion}
This work introduces a geometry-aware hybrid fusion (GA-HF) framework, the first
to address the challenges of sparse fisheye-LiDAR fusion for cost-effective
autonomous systems. By decoupling visual feature extraction from metric object
detection, processing fisheye data in a native polar grid to mitigate
distortion and LiDAR in Cartesian space for metric accuracy, we preserve
critical spatial density often lost in Cartesian representations. Our DA-WCM
effectively bridges the cross-modal quality gap, explicitly refining features
to suppress distortion artifacts in low-overlap regions.

Extensive experiments across three benchmarks, KITTI-360,
Dur360BEV and Fisheye3DOD, demonstrate that
GA-HF consistently achieves state-of-the-art detection quality. The
cross-dataset evaluation reveals that Cartesian BEV fusion suffers catastrophic
orientation degradation (mAOE~$>$~1.0) on real-world fisheye data, while our
hybrid design maintains robust performance even under imprecise calibration.
These results confirm that camera features are essential for semantic and
orientation estimation but require targeted geometric refinement to maintain
regression stability. Future work will explore temporal fusion mechanisms to
address velocity estimation and investigate end-to-end calibration refinement
to further improve fusion under uncertain extrinsics.

\bibliographystyle{IEEEtran}
\bibliography{icme2026references}

\end{document}